\documentclass[sn-chicago,iicol]{sn-jnl}

\jyear{2023}%

\theoremstyle{thmstyleone}%
\theoremstyle{thmstyletwo}%

\theoremstyle{thmstylethree}%

\usepackage[normalem]{ulem} 
\usepackage{amssymb}
\usepackage{amsmath}
\usepackage{latexsym}

\usepackage{url}
\usepackage{colortbl} 
\usepackage{xcolor}
\usepackage{balance}
\usepackage{hyperref}
\hypersetup{
    colorlinks=true,
    urlcolor=red,
}
\usepackage{ulem} 
\usepackage{adjustbox}

\definecolor{newcolor}{rgb}{.8,.349,.1}

\makeatletter
\def\blfootnote{\xdef\@thefnmark{}\@footnotetext}
\makeatother

\begin{document}

\title[Article Title]{Image-based Navigation in Real-World Environments via Multiple Mid-level Representations: Fusion Models, Benchmark and Efficient Evaluation\textsuperscript{\textsection}}

\author*[1,3]{\fnm{Marco} \sur{Rosano}}\email{marco.rosano@unict.it}

\author[1,5]{\fnm{Antonino} \sur{Furnari}}\email{furnari@dmi.unict.it}

\author[3]{\fnm{Luigi} \sur{Gulino}}\email{luigi.gulino@orangedev.it}

\author[2]{\fnm{Corrado} \sur{Santoro}}\email{santoro@dmi.unict.it}

\author[1,4,5]{\fnm{Giovanni Maria} \sur{Farinella}}\email{gfarinella@dmi.unict.it}

\affil[1]{\orgdiv{FPV@IPLAB - Department of Mathematics and Computer Science}, \orgname{University of Catania}, \orgaddress{\city{Catania}, \country{Italy}}}

\affil[2]{\orgdiv{Robotics Laboratory, - Department of Mathematics and Computer Science}, \orgname{University of Catania}, \orgaddress{\city{Catania}, \country{Italy}}}

\affil[3]{\orgdiv{OrangeDev s.r.l.}, \orgaddress{\city{Firenze}, \country{Italy}}}

\affil[4]{\orgdiv{Cognitive Robotics and Social Sensing Laboratory}, \orgname{ICAR-CNR}, \orgaddress{\city{Palermo}, \country{Italy}}}

\affil[5]{\orgdiv{Next Vision s.r.l.}, \orgaddress{\city{Catania}, \country{Italy}}}


\abstract{Robot visual navigation is a relevant research topic. Current deep navigation models conveniently learn the navigation policies in simulation, given the large amount of experience they need to collect.
Unfortunately, the resulting models show a limited generalization ability when deployed
in the real world.
In this work we explore solutions to facilitate the development of visual navigation policies trained in simulation that can be successfully transferred in the real world. 
We first propose an efficient evaluation tool to reproduce realistic navigation episodes in simulation.
We then investigate a variety of deep fusion architectures to combine a set of mid-level representations, with the aim of finding the best merge strategy that maximize the real world performances.
Our experiments, performed both in simulation and on a robotic platform, show the effectiveness of the considered mid-level representations-based models and confirm the reliability of the evaluation tool. The 3D models of the environment and the code of the validation tool are publicly available at the following link:~\href{https://iplab.dmi.unict.it/EmbodiedVN/}{https://iplab.dmi.unict.it/EmbodiedVN/}}




\maketitle

\blfootnote{\textsuperscript{\textsection} Accepted for publication in Autonomous Robots}

\section{Introduction}
\label{sec:introduction}
Creating a robot able to navigate autonomously inside an indoor environment relying just on egocentric visual observations is a challenging yet attractive research goal.
In recent advanced robotics applications, the visual data collected by the autonomous agent is generally processed by Deep Learning (DL) models to extract the properties of the environment in a more explicit form (e.g. by detecting the presence of objects, the presence of free space, the room type, the scene depth, etc.)~\citep{obj_det_survey, sem_seg_survey,taskonomy2018}, which can be eventually leveraged to perform operations in real-world scenarios~\citep{bonin2008visual,rescue_robots}.
Visual navigation approaches have been successfully applied when the goal to be reached is specified as coordinates~\citep{habitat19iccv}, images~\citep{zhu2017target}, object categories ~\citep{chaplot2020object}, room type~\citep{roomNav2020} and language instructions~\citep{chen2011language_nav}, showing that DL models are suitable tools to obtain robust navigation policies, given their ability to learn directly from data.
In particular, Deep Reinforcement Learning (DRL) showed that robotic agents can learn effective navigation policies from experience, by performing navigation episodes inside realistic simulators following a trial-and-error setup~\citep{zhu2017target, habitat19iccv}, and avoiding the need for densely annotated data, typical of classic supervised learning.
Unfortunately, despite the improved photo-realism of the simulated environments, navigation models trained in simulation struggle to effectively transfer to real spaces~\citep{robustPoliciesChen2020, rosano2020navigation}, due to different factors such as the visual discrepancy between virtual and real observations (domain shift) and the difference in robot dynamics between simulated and real world (i.e. real sensor measurements and robot movements are noisy and subject to failures). To address these limitations, several domain adaptation techniques have been proposed~\citep{wang2018da_survey} in order to reduce the gap between the two domains, usually by applying pixel or feature level transformations to the input images. 
In the case of pixel-level transformations, the goal is to translate images from the source domain to the target domain in order to make them visually indistinguishable~\citep{cyclegan,bousmalis2017unsupervised}; in the case of feature-level transformations, the visual encoder, usually a Convolutional Neural Network (CNN), is trained to map the representation vectors of images belonging to the two domains in the same compact subspace~\citep{adda,kouw2016feature}.

Other approaches~\citep{taskonomy2018} aim at extracting explicit, domain-invariant scene information from RGB observations, such as surface normals, keypoints and depth maps.
These so-called mid-level representations of the scene can model the visual structure of the space and have been proved useful to improve the performances of visual navigation models~\citep{navigateMidlevel2018}, and reduce the domain gap~\citep{robustPoliciesChen2020} between virtual and real world observations.
Despite all the collected evidences about their effectiveness, a systematic investigation on the impact of using multiple mid-level representations, with the aim of maximizing the transferability of PointGoal visual navigation models in the real world, has not been carried out yet. 
Intuitively, depending on what the agent observes during a navigation episode, some perception abilities could be more useful than others to successfully accomplish the navigation task (e.g. a depth map can be very useful in the presence of numerous obstacles in the scene, surface normals can help to detect a hidden step on the floor, etc.).
As we show in our experiments, providing the navigation models with a rich visual input and giving them the ability to adaptively weight the contribution of each mid-level representation depending on the agent’s perception, can lead to superior performances.

\begin{figure*}[t]
    \centering
    \includegraphics[width=0.99\linewidth]{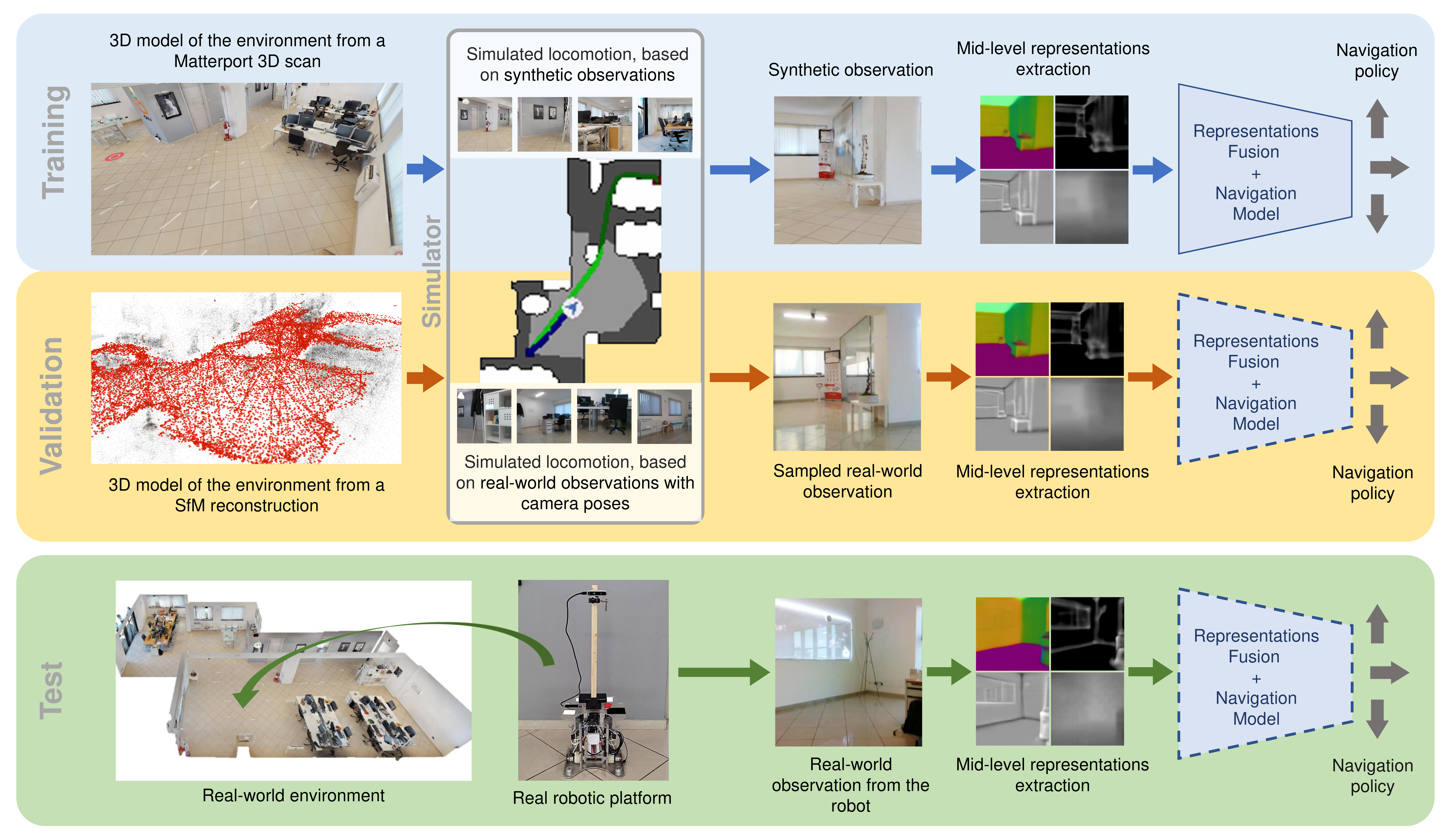}
    \caption{Illustration of the considered pipeline to train, evaluate and test visual navigation models. Training and evaluation are performed leveraging a locomotion simulator on two separate 3D models of the environment. During training (first row in light blue) a geometrically accurate 3D model of the environment is used to learn an optimal navigation policy using RL. To evaluate the navigation policy on a realistic scenario (second row in orange), a set of real-world images of the environment paired with camera poses is used to produce realistic navigation episodes in simulation.
    In both cases, a set of mid-level representations are extracted from the RGB observations, which are then adaptively combined by the proposed mid-level representations fusion architectures to perform the navigation task. The navigation models are then tested on a real robot to perform real-world navigation episodes (third row in green)}
    \label{figure: main}
\end{figure*}

Furthermore, as already highlighted in several related works~\citep{Wijmans2020DD-PPO:,gupta_cognitive}, the physical evaluation of visual navigation models in the real world remains difficult to carry out, mainly due to time and resources constraints (e.g., in terms of the human supervision required to perform the experiments) and the fragile nature of the robotic platforms. Indeed, hardware components, such as motors and batteries, are subject to wear and failures, whereas collisions with obstacles and bumpy rides can easily harm the integrity of the robotic platform. These limitations are a real obstacle preventing to carry out extensive evaluations in real scenarios.

In this work, we investigate both problems: 1) how to efficiently evaluate the performances of visual navigation models on realistic navigation episodes based on real observations, avoiding the physical deployment to a robotic platform; 2) how to train visual navigation models entirely in simulation, that can be successfully deployed in the real world.
To this end, we proposed an evaluation tool, built on top of the Habitat simulator~\citep{habitat19iccv}, to efficiently assess the performance of navigation models in a realistic setting, while avoiding the time-consuming evaluation on a real robot. The tool leverages two 3D models of the same environment: a geometrically accurate 3D model to generate the virtual navigation episodes in simulation, and a sparse 3D model consisting of a set of real-world images attached with their camera poses, used to provide the generated navigation episodes with real observations.

The evaluation tool has been then employed to investigate whether mid-level representations can improve the transfer of a policy learned in simulation to the real world. To this aim, we exploited the DL models proposed in the work of~\cite{taskonomy2018} to extract mid-level representations from RGB observations and we considered a variety of deep learning architectures for visual navigation which perform early, mid and late fusion of the extracted representations.
More in details, four mid-level representations, namely surface normals, keypoints3D, curvature, depth, have been considered as the most prominent to perform PointGoal navigation, given their ability to capture the most important geometric properties of the environments. Figure~\ref{figure: main} contains examples of the examined mid-level representations (``Mid-level representations extraction'' step).

All the proposed fusion navigation models have been trained using the Habitat simulator~\citep{habitat19iccv} on the synthetic version of a real environment following a DRL setting, and have been evaluated on realistic virtual navigation episodes based on real-world observations using the proposed evaluation tool. 
To confirm the effectiveness of our tool we also tested the navigation models in the real environment using a custom robotic platform, equipped with accurate actuators.
A complete overview of the proposed framework is depicted in Figure~\ref{figure: main}.
Overall, all the proposed navigation models showed a good behavior when deployed in the real world, supporting the performance estimated by our evaluation tool, albeit reporting different navigation capabilities in the various navigation episodes.

We observed that the proposed representations fusion models are effective at reducing the real and simulated observations gap, reaching comparable performances with navigation models that used real-world images to perform adaptation. Also, using multiple mid-level representations as input resulted in better performing navigation policies, even in the case of simple fusion architectures, with more advanced fusion strategies reporting the best results among all fusion models.

In summary, the contributions of this work are as follows:
\begin{enumerate}
\item we proposed an evaluation tool, built on top of the Habitat simulator~\citep{habitat19iccv}, to reproduce realistic navigation trajectories in simulation by leveraging a geometrically accurate 3D model of an indoor environment and a set of real-world images paired with the camera poses, sampled from a Structure from Motion (SfM)~\citep{colmap} reconstruction of the same environment. We showed the effectiveness of the evaluation tool by performing a test session on real-world trajectories using a robotic platform. Our tool allows a fast and inexpensive assessment of the models capabilities and represents a good proxy for estimating real world performance.
\item we proposed a variety of learning-based visual navigation models performing mid-level representations fusion, to learn optimal navigation policies in simulation which can be directly deployed in the real world, without performing any additional domain adaptation. Each model receives a variable amount of mid-level representations and follows a different combination strategy, to learn how to adaptively balance the contribution of the representations and maximize the final performances;
\item following an extensive evaluation both in simulation and on a real robotic platform, we showed how the number of used mid-level representations and the type of adopted fusion architecture can impact the final performance of navigation models. 
Overall, the navigation models benefited from multiple mid-level representations and showed comparable performances to models trained using real-world observations;

\end{enumerate}

The remainder of the paper is organized as follows. Section~\ref{sec:relatedworks} discusses the related works. In Section~\ref{sec:method}, we describe the proposed approach. The experimental settings are discussed in Section~\ref{sec:experimentalsettings}, whereas the results are presented in Section~\ref{sec:results}. Section~\ref{sec:conclusion} concludes the paper and gives hints for future works.

\section{Related Works}
\label{sec:relatedworks}
Our work relates to approaches which belong to a range of topics, including simulators for visual navigation, embodied visual navigation, simulated to real domain adaptation and visual representation. We report the most relevant connections to our work with respect to the state-of-art in the subsections below.

\subsection{Embodied Navigation Simulators}
The development of advanced simulators~\citep{savva2017minos,house3d,xiazamirhe2018gibsonenv,nvidia2021isaac} used in conjuction with realistic large-scale 3D indoor datasets laid the foundations for the design of learning-based navigation models which can learn the desired behaviour through realistic interactions with the scene.
To foster the research of robotic agents that perform increasingly complex tasks, more recent works~\citep{newhabitat2021,gibson2.0} released highly interactive environments comprised of a large number of active elements (e.g. moving pedestrians, interactive furniture, etc.), within which the agent can widely experiment a large variety of realistic interactions.
Embodied simulators are designed to be used with third-party 3D datasets, which have different characteristics and are designed following different approaches.
The 3D spaces proposed in the works of~\cite{Matterport3D,xiazamirhe2018gibsonenv,dai2017scannet} reproduce real-world indoor rooms and have been acquired using dedicated 3D scanners. This allows for the collection of a large number of photo-realistic 3D environments at a relatively low cost, but the final 3D reconstruction may present holes or artifacts due to imperfect scans. In contrast, the 3D models proposed by~\cite{replica19arxiv,openRooms2021,kolve2017ai2,newhabitat2021} represent replicas of realistic indoor spaces, accurately designed by artists.
The survey of~\cite{moller} contains a detailed section on state-of-art datasets and simulators for robot navigation.

To facilitate the assessment of navigation performance in the real-world,~\cite{robothor} released a set of 3D virtual environments for training purposes and allowed researchers to physically test the obtained navigation models on the real equivalents through a remote deployment application.

In this work, we aim to train visual navigation models in simulation that can be directly deployed in the real world. Given the need to efficiently assess their navigation capabilities, we extended the functionality of the Habitat simulator~\citep{habitat19iccv} to reproduce realistic navigation episodes containing real-world observations. The proposed framework allows for a good estimation of the real-world performances, while avoiding the deployment of the policy to a physical robotic platform.

\subsection{Embodied Visual Navigation}
The problem of robot visual navigation has been studied for decades by the research community~\citep{bonin2008visual,thrun2002probabilistic}. In its classic formulation, the navigation process can be thought as a composition of sub-problems: 1) construction of the map of the environment; 2) localization inside the map; 3) path-planning to the goal position; 4) execution of the navigation policy.
The environmental map can be provided beforehand or reconstructed with a SfM pipeline~\citep{colmap} using a set of images of the space. The localization is then performed by comparing new observations with the previously collected set of data.
Also, in SLAM-based methods~\citep{cadena2016past,fuentes2015visual}, the reconstruction of the map and the localization tasks are performed at the same time.
The navigation is then performed after a path to the goal is computed. 
These methods have been implemented in several scenarios but they present significant limitations, such as the limited scalability to large environments, the accumulation of the localization error and the limited robustness to dynamic scenarios.
Recently, learning-based visual navigation approaches emerged as effective alternatives to the classic navigation pipelines, promising to learn navigation policies in a end-to-end way, receiving images as input and returning actions as output, avoiding to explicitly model all the intermediate steps~\citep{zhu2017target,mirowski2016learning}.
Depending on the type of goal, deep learning models can perform ObjectGoal~\citep{chaplot2020object,morad2021embodied} or RoomGoal~\citep{roomNav2020} navigation, follow instructions expressed in natural language~\citep{chen2011language_nav,krantz2020navgraph,anderson2018vision,fried2018speaker} or act to answer questions about properties of the environment~\citep{das2018embodied,gordon2018iqa}.

When goals are specified as a coordinates (PointGoal) or observations of the environment (ImageGoal), the task is referred to as \textit{geometric navigation}, given the requirement of the navigation model to reason about the geometry of the 3D space in order to accomplish the task.
Recent geometric navigation approaches investigated the use of a variety of learning architectures: ~\cite{zhu2017target} used reactive feed-forward networks for ImageGoal navigation;~\cite{habitat19iccv} included a recurrent module to embed the past experience, enforcing the sequential nature of navigation;~\cite{Wijmans2020DD-PPO:} improved the scalability of the model collecting billions of frames of experience.~\cite{chen2020soundspaces} introduced the use of sounds together with images to reason about the surrounding space and to guide the agent towards the goal.~\cite{chaplot2020learning} and~\cite{chen2018slam} used spatial memories and planning modules, whereas~\cite{savinov2018semiparametric} and~\cite{chaplot2020topological} used topological memories to represent the environment.

The methods investigated in this paper fall in the class of geometric navigation approaches, where the model needs to reach a goal specified as coordinates. Similarly to the method proposed by~\cite{habitat19iccv}, we trained a set of RL-based navigation models consisting of both convolutional and recurrent modules. 
However, our approach differs for the type of input the navigation models receive and for the final objective.

\subsection{Simulated to Real Domain Adaptation}
Domain adaptation methods applied to visual navigation aim to learn domain-invariant representations for virtual and real observations which present difference in style although depicting the same content. Existing simulators for visual navigation generally adopt strategies to limit this gap, by following two strategies: by making the environment highly photo-realistic or by randomizing the properties of the virtual environment. In the former approach, the goal is to make the simulation appear as similar to the real world as possible. In the latter, the idea is to make the model experience a highly dynamic environment to avoid overfitting to a specific style and to allow for a style-agnostic representation learning of the space. Domain randomization was successfully applied in robotic grasping~\citep{james2019grasp}, drone control~\citep{Loquercio2020DeepDR,sadeghi2016cad2rl}, vision-and-language navigation~\citep{vln-pano2real}.
When the domain gap persists, specific strategies should be adopted. For example, when real-world data is available beforehand, it is possible to train the navigation models on synthetic data and then fine-tune them on real observations~\citep{grasp2017finetune,rosano2020navigation}. Real observations can also be employed to perform adaptation at the feature-level~\citep{adda,kouw2016feature}, pixel-level~\citep{hu2018duplex,cyclegan} or at both levels~\citep{hoffman2018cycada}.
More recently, different works were proposed to address specifically the transfer of visuomotor policies from the simulated to the real world. For instance,~\cite{lichaplot2020unsupervised} proposed a GAN-based model to decouple style and content of visual observations and introduced a consistency loss term to enforce a style-invariant image representation.~\cite{rao2020rl} introduced a RL-aware consistency term to help preserving task-relevant features during image translation.~\cite{truong2021bidirectional} followed instead a bi-directional strategy, using a CycleGAN-based~\citep{cyclegan} real to simulated adaptation model for the visual observations and a simulated to real adaptation module for the physical dynamics.
Rather than using adaptation modules to reduce the gap between real and simulated observations, we followed the idea of~\cite{navigateMidlevel2018} and~\cite{robustPoliciesChen2020} and trained our navigation policy on top of mid-level representations, which contain crucial geometric and semantic cues of the environment and are invariant to the navigation scenario.
The idea of combining different visual representations to improve the navigation abilities of an agent has also been explored in other approaches~\citep{mousavian,morad2021embodied} but their fusion mechanisms have often been limited to a simple stacking of the different input representations.
More recently,~\cite{situational2019} explored this opportunity by proposing a set of DL-based fusion architectures in the context of ObjectGoal navigation, but their experiments were conducted in a poorly realistic setup (a discretized grid-world) and did not consider the deployment of the learned models in the real world.
In contrast, our approach aims at learning how to leverage the correct combination of visual geometric cues that better transfer to the real world, considering a continuous state space and a goal specified as coordinates.

\subsection{Evaluation of Visual Navigation Systems}
Most works on visual navigation systems for robot navigation focused on the advancement of control systems, with all experiments carried out exclusively on simulators, leaving the domain adaptation problem for further investigations~\citep{mirowski2016learning,chen2018slam,Wijmans2020DD-PPO:}. Some studies considered the gap between simulation and real world, eventually performing qualitative tests on real-world trajectories using custom robotic platforms~\citep{zhu2017target,chaplot2020learning}.
For instance,~\cite{arewemakingprogress} conducted a study to measure if advancements recorded in simulation reflect advancements in the real world. Given that no special strategies were adopted to reduce the visual domain gap, the authors set up a perfect replica of the room available in simulation to perform the evaluation of the visual-based navigation policies using a physical robot.
To minimize avoid the domain gap,~\cite{tai2017_rl_realrobot} trained a RL-based navigation policy in simulation using a combination of Lidar signals and Depth images. The evaluation, performed directly on a real robot, confirmed the importance of using domain-invariant representations as input to the navigation models, in order to avoid additional adaptation steps.
Conversely, in their attempt to design a domain-agnostic navigation system for flying drones,~\cite{sadeghi2016cad2rl} showed that deploying a navigation policy trained in simulation on a real drone is challenging and that the performance obtained from the evaluation on more photorealistic environments does not directly reflect real-world performance.
In a similar way, classic navigation approaches rely on simulators for their development and evaluation, even if they do not require a training procedure~\citep{takaya2016simulation,wang2017autonomous,collins2021review}. These simulators are usually able to reproduce the physical dynamics of the real world though lacking in photorealism~\citep{koenig2004design,carpin2007usarsim}. 
In contrast with the aforementioned methods, our evaluation tool aims to ease the assessment of the performance of visual-based navigation policies on realistic observations directly in simulation, streamlining the procedure required to deploy the navigation system on a real robot.

\section{Method}
\label{sec:method}
In this section we first provide details about the framework used to train and evaluate our navigation policies, including information on the 3D models reconstruction and alignment pipeline and on the generation of the realistic navigation episodes. We then proceed by describing the navigation problem setup in presence of multiple mid-level representations and the type of visual representations fusion strategies that have been adopted.
\begin{figure}[t]
    \centering
    \includegraphics[width=0.99\linewidth]{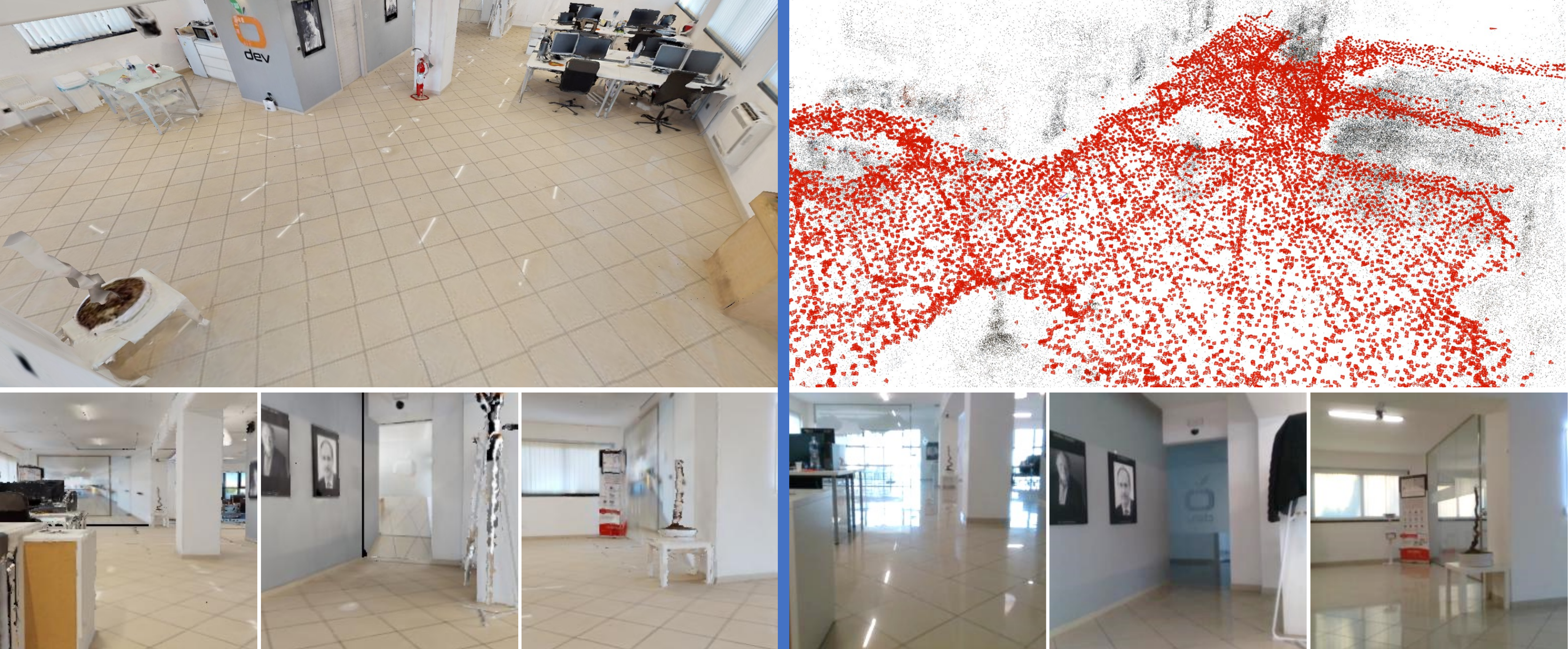}
    \caption{On the left, a view of the geometrically accurate 3D model of the considered office environment. It allows the sampling of images from any position but it is limited in terms of photo-realism. On the right, a view of the 3D models of the same indoor environment, reconstructed following a Structure from Motion (SfM)~\citep{colmap} pipeline. It contains a sparser collection of real-world images paired with their camera poses (each red marker represents the position of an image in the 3D model)
    }
    \label{figure: virtual_real_3d_models}
\end{figure}
\subsection{3D Models Reconstruction and Alignment}
\label{subsec:acquisitionorientation}
Our training and evaluation tool requires the acquisition of two 3D models of the same environment: a geometrically accurate 3D model that can be acquired using a 3D scanner, such as Matterport 3D\footnote{\url{https://matterport.com/cameras/pro2-3D-camera}}, and a photo-realistic 3D model reconstructed from a set of real-world observations, using a SfM algorithm~\citep{colmap}.
The first model is an accurate replica of the real environment but with limited photorealism. The scanning process returns a 3D mesh that can be natively imported inside Habitat~\citep{habitat19iccv} and used to train the navigation policy.
On the contrary, the second model is a sparse photo-realistic but geometrically inaccurate reconstruction of the environment. The SfM process returns a 3D pointcloud in which all images are labeled with their camera pose (position and orientation).
Figure~\ref{figure: virtual_real_3d_models} compares the two 3D models.
It is worth noting that this model can not be directly used in the Habitat simulator and a dedicated interface was developed as part of our tool to allow its use inside the simulation platform, as described in Section~\ref{subsec:realepisodegeneration}.
Because the two 3D models are acquired separately using two different approaches, they might present a scale and a rotation offset, that should be minimized by following a maps alignment procedure.
One possible solution is to manually search the parameters of the affine transformation, that is then applied to one or both 3D models to match the coordinates of the other 3D model. To make this process automatic, we leveraged an image-based alignment procedure\footnote{We used the \textit{model\_aligner} function of the COLMAP software~\url{https://colmap.github.io/faq.html}} to transform the coordinate system of the 3D model containing real-world observations to match the one of a set of observations sampled from the geometrically accurate 3D model. To this end, we used the Habitat simulator to collect images from random locations together with their camera pose. Although the images belong to two different 3D models and their appearance does not match perfectly, the aligning procedure turned out to be robust against visual differences and succeeded with the coordinate system transformation.

\subsection{Generation of Realistic Navigation Episodes in Simulation}
\label{subsec:realepisodegeneration}
Once the 3D models are aligned, they can be exploited to generate realistic navigation episodes in simulation. At first, the navigation trajectory is generated by the simulator on top of the geometrically accurate 3D model. Then, the virtual agent performs the navigation task and at each step the perceived virtual observation is systematically replaced with the real-world image which is closest in space to the current agent position. In more detail, at each step, the current pose of the agent is extracted from the simulator and it is used to retrieve the nearest real image from the 3D model containing real-world observations.
The retrieved real observation is then processed by the visual navigation module in order to predict the action to take in the virtual environment. This process is repeated until the end of the navigation episode.
Considering that the agent moves on the floor surface and that its camera does not change its height nor its pitch and roll angles, the 6DoF camera poses were transformed to 3DoF coordinates, with the first two degrees of freedom representing the X and Z cartesian coordinates on the ground plane and the third degree of freedom representing the camera orientation as the angle along the Y axis, perpendicular to the XZ plane. 
We transformed the cameras heading angle $\theta$ to unit vectors $(u,v)$, where $u=cos\theta$ and $v=sin\theta$, and calculated the angle difference as the cosine similarity between the corresponding vectors. In our experiments we found that a similarity threshold of 0.96 ensures good results.
After filtering the real-world images by angle, we apply a second filter to the resulting subset of images based on the X-Z coordinates. Finally, the nearest image is chosen to replace the virtual observation.
Because the image retrieval time is crucial to perform a fast policy evaluation, we leveraged the efficiency of the FAISS library~\citep{FAISS} to perform a fast search on a large set of thousands of records in a fraction of a second.
As a result, the navigation episode is performed in simulation but the policy is obtained by processing real-world observations.

\begin{figure*}[t]
    \centering
    \includegraphics[width=0.99\linewidth]{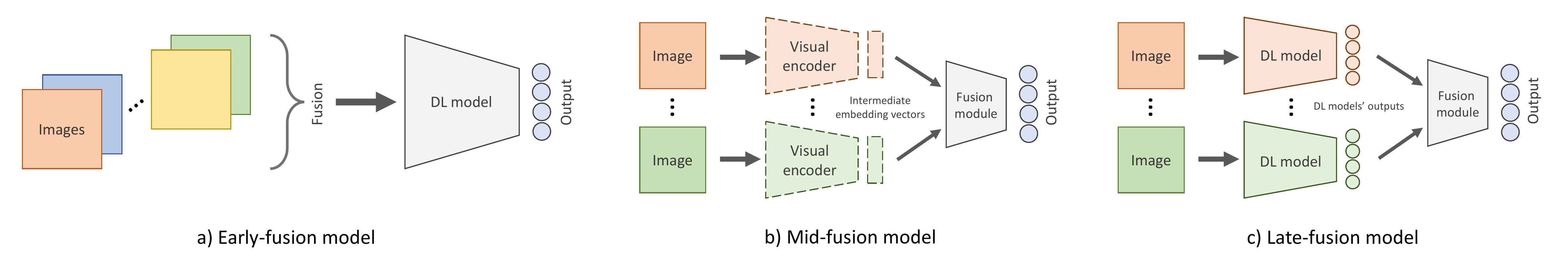}
    \caption{Popular Deep Learning models used to perform visual representations fusion. a) In Early-fusion models, the representations are combined at an early stage and are then provided as input to the deep model. b) In the case of Mid-fusion models, each representation is processed by a separate encoder, that outputs an intermediate embedding vector (dashed line rectangles). Therefore, all vectors are combined by a fusion module to produce the model's final output. c) Differently from the previous two, the Late-fusion model is an ensemble of indipendent models, each one processing a single representation and producing a distinct output (e.g. the action to perform). A fusion module then collects the outputs of the models to return the final decision. In the figure, the final models' output are denoted by blue circles}
    \label{figure: fusions}
\end{figure*}

\subsection{Navigation Problem Setup}
We consider the PointGoal visual navigation task in indoor environments. In this context, an agent equipped with an RGB camera is placed at a random location of the environment and is required to navigate towards a goal location, indicated through a set of coordinates, relying solely on the visual observations to reason about the surrounding space and execute the best possible actions. No explicit information about the layout of the environment is provided to the agent.
At each timestep $t$, the agent receives a RGB observation $o_{t}$ that is processed by a set of transformation models $f_{1}, f_{2},...,f_{n}$ (we use the models provided by~\cite{taskonomy2018}) to output a list of mid-level representations $m_{t}^{1}, m_{t}^{2}, ..., m_{t}^{n}$.
From Figure~\ref{figure: main} it is possible to observe examples of mid-level representations obtained from the respective RGB images and the different scenes properties they are able to capture. These representations are then passed to a fusion module which learns how to combine them in order to produce a final compact vector containing the most meaningful information about the agent's current observation. Our navigation policy is parametrized by a neural network $\pi(a_{t}\mid m_{t}^{1},...,m_{t}^{n},g)$ which, given the visual representations $m_{i}^{j}$ and information about the goal to reach $g$, predicts the action $a_{t}$ to perform at time $t$. This process is repeated until the goal or a given steps budget is reached.

The navigation models were trained entirely in simulation following a RL setup. In RL, the agent performs actions inside the virtual environment and collects rewards or penalties (negative rewards), depending in whether the actions led to reduce the distance to the goal or not. The objective of the training process is to find an optimal navigation policy $\pi^{*}$ which allows the agent to find the shortest path to the goal by maximizing the sum of the collected rewards.

\subsection{Mid-level Representations Fusion}
\label{subsec:fusionmethods}
In this work, we leverage the mid-level representations proposed by~\cite{taskonomy2018}, able to capture a variety of different geometric and semantic properties of the observed environment. 
To investigate the benefits that a visual representations fusion strategy can offer to models performing PointGoal navigation, we proposed a variety of deep convolutional networks to perform early, mid and late fusion. Figure~\ref{figure: fusions} shows an overview of the investigated fusion schemes. More specifically, we considered five different visual encoders:
\begin{itemize}
    \item a classic convolutional model performing early-fusion of the mid-level representations. It represents the most simple combination strategy and can be considered as a baseline for more elaborated fusion models (Figure~\ref{figure: fusions}a);
    \item two convolutional models with a channel-level attention mechanism. This architecture represent a variant of the Early-fusion model depicted in Figure~\ref{figure: fusions}a, which performs a weighting of the feature maps after every convolutional layer, similarly to what is done with the Squeeze-and-excitation networks~\citep{squeeze_excitation}. Assuming that different feature maps contain different properties of the input observation, this architecture can learn to focus on the most relevant ones. The two models differ for the type of layer pooling used in the attention branches; Section~\ref{subsec:proposedmodels};
    \item a Mid-fusion model (Figure~\ref{figure: fusions}b) which processes each mid-level representation in a dedicated convolutional branch, to then aggregate their outputs in the final shared layers. This architecture can specialize portions of the network to exploit the visual cues contained in specific mid-level representations;
    \item a Late-fusion model (Figure~\ref{figure: fusions}c) which represents an ensemble of networks, each of them trained separately on a single mid-level representation. Each network outputs the probability of taking an action and a final policy fusion module aggregates them to select the final action, based on a context summary representation.
\end{itemize}

Each visual encoder is responsible of processing the visual mid-level representations and is followed by a controller, which receives the output of the visual encoder and further useful data to output the navigation policy.
More detailed information about the architectures of the proposed fusion models is reported in Section~\ref{subsec:proposedmodels} and in Figure~\ref{figure: proposed_models}.

\section{Experimental Settings}
\label{sec:experimentalsettings}
\subsection{Dataset Acquisition}
We carried out our experiments in an office environment of about $150$ square meters.
The geometrically accurate 3D model was acquired using a Matterport 3D scanner and the resulting 3D mesh was imported inside the Habitat simulator to perform the training of the navigation policies. Instead, the 3D model containing real-world observations was reconstructed for testing purposes using the COLMAP~\citep{colmap} software, starting from a set of ~32k RGB images of the environment, collected using a robotic platform equipped with a Realsense d435i camera\footnote{\url{https://www.intelrealsense.com/depth-camera-d435i/}}. 
This resulted in a sparse 3D pointcloud where each image is labeled with its camera pose relative to the 3D reconstruction.
To capture the real-world images, the robotic agent followed a simple exploration policy aimed at covering all the traversable space as more uniformly as possible, proceeding along straight trajectories, stopping and turning around by a random angle to avoid collisions and continue the acquisition. This procedure was carried out automatically by the robot, thanks its on-board infrared (IR) sensors that can perform short-range obstacle detection. The real-world image set was acquired in about 3.5 hours at 3fps, with a robot's maximum speed of 0.25m/s.
As already mentioned in Section~\ref{subsec:acquisitionorientation}, the 3D model containing real-world observations was aligned to match the coordinate system of the geometrically accurate 3D model. For this purpose, an ``alignment set'' of 6k images was randomly sampled from the virtual environment together with the relative camera poses. These images were registered inside the 3D model using COLMAP and then used by the image-based alignment function to perform the final match of the coordinate system.
\begin{figure*}[!t]
    \centering
    \includegraphics[width=0.8\linewidth]{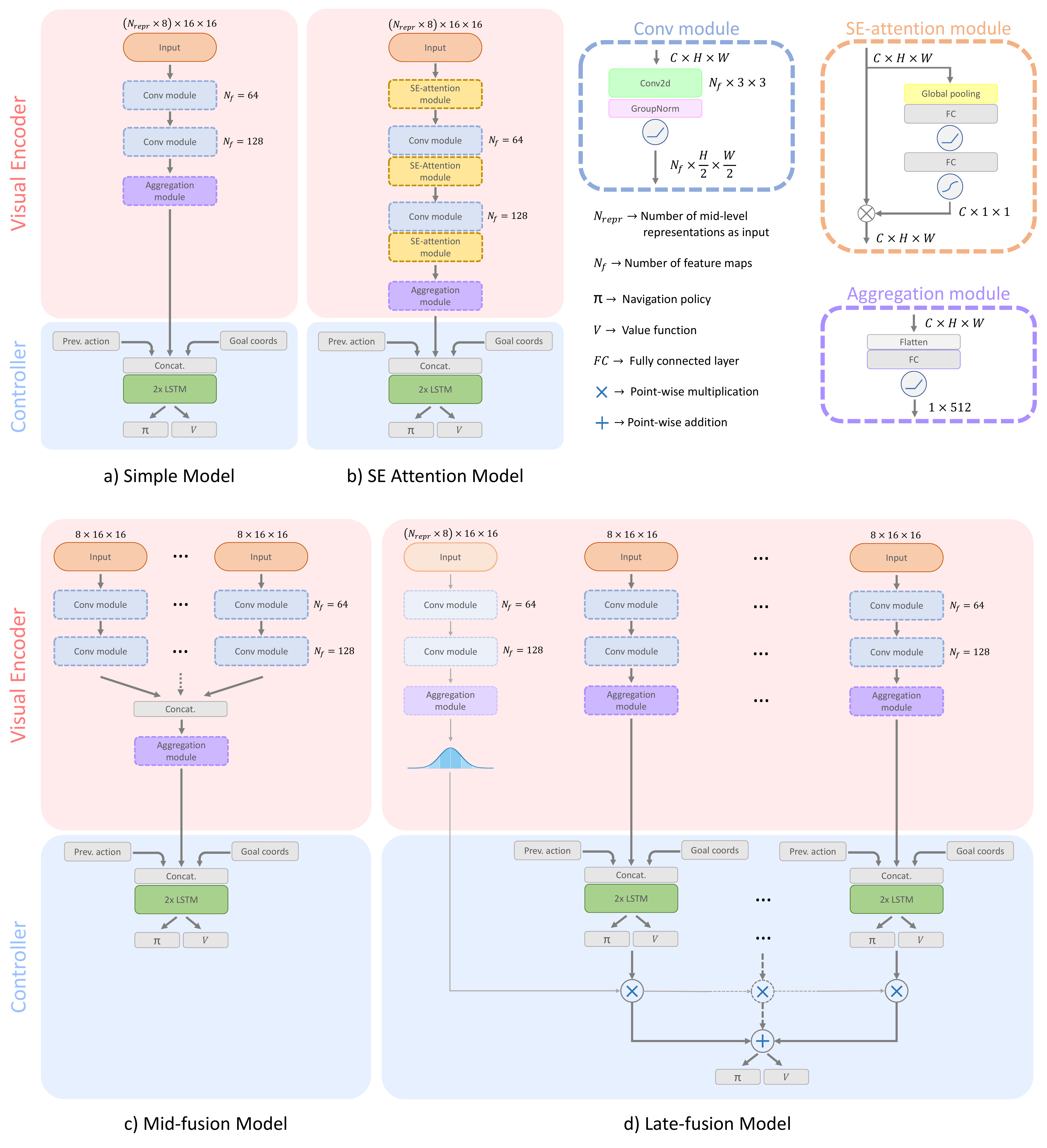}
    \caption{Overview of the proposed visual navigation models, which follow distinct mid-level representations fusion strategies. All the models are comprised of two parts: 1) a visual encoder and 2) a controller. The visual encoder (red background) is responsible for effectively combining the different mid-level representations provided as input to produce a meaningful vector embedding of the scene. The controller (light blue background) takes this embedding as input, together with additional information about the coordinates of the navigation goal and the previously performed action, to output the action to take and the estimated ``quality'' of the current state to reach the destination. The LSTM layers allow the model to embed the history of the navigation episode at each timestep, given the sequential nature of the task. Each model was decomposed in modules, which are detailed on the top right of the figure. See the text for the discussion of the different models}
    \label{figure: proposed_models}
\end{figure*}
\subsection{Proposed Navigation Models}
\label{subsec:proposedmodels}
In our experiments we leveraged four mid-level models from~\cite{taskonomy2018} to extract \textit{surface normals, 3D keypoints, curvature, depth map}.
We considered these representations because they are able to capture different geometric properties of the environment, which is ideal given that in our navigation setup the goal is specified as coordinates in the space and that the task requires geometric reasoning.
Each of these models receives a $256\times256$ RGB image and outputs a compact tensor of size $8\times16\times16$.
We found that this compact representation provides the navigation model with the information required to successfully perform the downstream task. Moreover, it allows the design of compact navigation models which results in faster training and easier deploy in real robotic platforms, whose computational resources are usually limited.
Specifically, we proposed five navigation models implementing five distinct visual encoders, also depicted in Figure~\ref{figure: proposed_models}:
\begin{itemize}
    \item the \textit{``Simple"} model consists of two convolutional layers with $3\times3$ kernels with 64 and 128 intermediate feature maps respectively. After each convolutional layer, we introduced a GroupNorm normalization layer to take into account the highly correlated data in the batch and a ReLU activation function. The aggregation module collects all the considered mid-level representations and stacks them along their features dimension to produce a unified representation of the agent's observation. This representation is then provided as input to the model. Figure~\ref{figure: proposed_models}(a) illustrates the architecture of the model;
    \item the \textit{``Squeeze-and-Excitation"} (SE) model introduces a feature-level attention module after each convolutional layer, including the input layer, to weigh the different feature maps depending on their content. Following~\cite{squeeze_excitation}, each attention module consists of a global pooling layer, two fully connected (FC) layers separated by a ReLU activation function and a final sigmoid activation function which returns the weights to perform the feature map-level attention. We tested two variants of the same model, one with global average pooling and the other with global max pooling in the attention module. We refer to these variants as ``SE attention (avg pool) model'' and ``SE attention (max pool) model'' respectively. Figure~\ref{figure: proposed_models}(b) depicts the aforementioned model;
    \item the \textit{``Mid-fusion"} model consists of a number of parallel visual encoders equal to the number of input representations. In this architecture, each encoder has the chance to focus on a single mid-level representation and the final output of the model is given by the combination of the intermediate outputs of the various visual branches. In practice, each branch is represented by the same visual encoder of the ``Simple model", which produces a compressed visual representation as output. These ``intermediate'' representations coming out from all the branches are subsequently concatenated along their channel dimension to form the final visual representation. We also explored more advanced combination strategies but in our experiments the simple concatenation returned the best results. Figure~\ref{figure: proposed_models}(c) presents a scheme of the model;
    \item the \textit{``Late-fusion"} model differs from the previous models because it consists of a set of full navigation models (visual encoder + controller) pre-trained independently on single distinct mid-level representations. At each navigation step, the models output action candidates and they are combined together depending on the current agent's perception to output the final action. More specifically, each navigation model outputs a probability distribution over a discrete set of actions that the agent can perform and an additional policy fusion module is responsible to adaptively weigh the single model's outputs to obtain the final action probability. The policy fusion module follows the same architecture of the visual encoder of the ``Simple Model", which takes a stack of the considered mid-level representations as input and outputs the weights (a probability distribution over the number of models) to balance the contribution of every navigation model to the final output. To train the policy fusion module, all the navigation models were trained beforehand and then frozen.
    Figure~\ref{figure: proposed_models}d) summarizes the entire architecture.
    
    Given $n_{m}$ the number of considered models, $n_{a}$ the number of actions in the discrete action set, $\mathbf{A}\in\mathbb{R}^{ n_{m}\times n_{a} }$ the matrix containing the models' actions candidate vectors and $\mathbf{w}_{pfm}\in\mathbb{R}^{ n_{m}\times 1}$ the output of the policy fusion module, the final action probability distribution $\mathbf{y}\in\mathbb{R}^{ n_{a}\times 1}$ is equal to:
    \begin{gather*}
    \mathbf{y}=\mathbf{A}^{T}\mathbf{w}_{pfm}
    \end{gather*}
    
\end{itemize}
The controller consists of two LSTM layers~\citep{lstm} which take as input the visual representation coming from the visual encoder, the action produced by the navigation model at the previous timestep and the information about the goal coordinates relative to the robot's current position, and outputs an action probability distribution together with a value representing the ``quality'' of the current agent's location given the goal to be reached (i.e. it is an Actor-Critic RL model~\citep{actor-critic}). The vectors informing about the previous action and the goal coordinates are both projected to two separate vectors of size 32 and then concatenated to the output of the visual encoder, which is a 512-d vector, to produce the final 576-dimensional vector that is fed to the controller. The use of recurrent layers helps the model to deal with the sequential nature of the navigation task.

\subsection{Training Details and Evaluation}
All the proposed navigation models have been trained on the synthetic version of the considered office environment, following the setup of~\cite{Wijmans2020DD-PPO:}. We used the Habitat simulator~\citep{habitat19iccv} to sample a set of 100k virtual navigation episodes beforehand, which we used to train our navigation models for 5 million frames each. This threshold was set experimentally as it allowed the agents to collect the required experience and to converge to optimal training metrics.

The navigation models have been trained with one, two, three and four mid-level representations as input, all from scratch.

In order to speed-up the training procedure we leveraged the DD-PPO architecture proposed by~\cite{Wijmans2020DD-PPO:}, a distributed variant of the popular PPO reinforcement learning algorithm~\citep{schulman2017ppo}, which allows multiple agents to be trained in parallel on one or multiple GPUs. Additionally, we adopted an input caching system that resulted in a doubled training speed (from 80fps to 160fps for a model receiving two representations as input, trained with 4 parallel processes per gpu on two Nvidia Titan X).

At each timestep the agent chooses one of the four possible actions available: \textit{move straight by} $0.25m$, \textit{turn left by} $10^\circ$, \textit{turn right by} $10^\circ$, \textit{STOP}.
The navigation episode ends when the \textit{STOP} action is performed or when the maximum number of execution steps is reached. We fixed this threshold to 200 steps based on the size of the environment.

Our visual navigation models have been evaluated using two standard metrics to measure the performance of agents navigating indoor spaces: Success Rate (SR) and Success
weighted by (normalized inverse) Path Length (SPL).
The SR measures the effectiveness of the navigation policy at reaching the goal. It is defined as the ratio between the number of successful navigation episodes and the total number of performed episodes:
\begin{equation}
    \frac{1}{N}\sum_{i=1}^{N}{S_{i}}
\end{equation}
where $N$ is the total number of performed episodes and $S_i$ is a boolean value indicating whether the \textit{i-th} episode was successful.
The SPL takes into account the path followed by the agent and can be thought as a measure of efficiency of the navigation model with respect to an agent following the shortest geodesic path to the goal. The SPL is defined as:
\begin{equation}
    \frac{1}{N}\sum_{i=1}^{N}{S_{i}\frac{l_{i}}{max(l_{i}, p_{i})} }
\end{equation}
where $N$ is the total number of performed episodes, $S_i$ is a boolean value indicating the success of the \textit{i-th} episode, $l_i$ is the shortest geodesic path length from the starting position to the goal position of the \textit{i-th} episode and $p_i$ is the agent's path length in the \textit{i-th} episode. In case of a perfectly executed navigation episode, the SPL assumes the value of 1. On the contrary, if the navigation policy fails, it assumes the value of 0.
The navigation models have been evaluated in simulation on 1000 episodes defined by a starting and a goal positions. Episodes have been sampled beforehand taking into account their complexity. Indeed, to avoid excessively simple navigation episodes, they have been filtered to ensure that the ratio between the geodesic distance and the euclidean distance from the starting position to the goal is greater than $1.1$, as already suggested in the work of~\cite{habitat19iccv}.
An episode is considered successful if the agent calls the \textit{STOP} action within $0.20m$ from the goal, and unsuccessful if it is called beyond that distance or if the the maximum number of steps (200) is reached. For the evaluation we used the same steps budget per episode used during train.

\subsection{Baseline Navigation Models}
We compared the proposed models with three baselines, which share the same architecture but receive different types of RGB images as input. They consist of: 1) a SE~\citep{squeeze_excitation}-ResNeXt50~\citep{resnext}, a larger visual encoder compared to the ones used to process the mid-level representations, suitable to process the lower level information contained in the input images; 2) a controller, identical to the one used in the proposed mid-level models. We considered the DL model pretrained on the Gibson~\citep{xiazamirhe2018gibsonenv} and Matterport3d~\citep{Matterport3D} datasets for 2.5 billion steps, released by~\cite{Wijmans2020DD-PPO:}, that was then adapted to our office environment.
Specifically, the baselines models are as follows:
\begin{itemize}
    \item the \textit{``RGB Synthetic"} model was trained on the synthetic observations coming from the proposed virtual environment, for 5 million steps. It is considered to assess to what extend a navigation model trained purely in the virtual domain can be transferred directly to the real world, without further adaptation or supervision;
    \item the \textit{``RGB Synthetic + Real"} model was trained for 2.5 million steps on the synthetic observations and then fine-tuned for other 2.5 million steps on real-world images. The real observations used to fine-tune the navigation model belong to a different SfM reconstruction, which consists of ~25K new acquired images of the same indoor space. As for the 3D model used for evaluation, it was first aligned to the geometrically accurate 3D model and then used for training purpose. This navigation is chosen to investigate whether using observations of the target domain during training can improve the navigation performance, despite knowing that collecting and exploiting them could be often expensive or unfeasible. 
    We expect this model to reach a near-optimal navigation performance;
    \item the \textit{``RGB Synthetic + CycleGAN"}~\citep{cyclegan} model was trained for 2.5 million steps on the synthetic observations and then fine-tuned for other 2.5 million steps on ``fake'' real observations, obtained by transforming the synthetic images to look like the real ones using CycleGAN~\citep{cyclegan}, a pixel-level unsupervised domain adaptation model trained on two sets of unpaired synthetic and real-world images (5K for each domain), randomly sampled from the virtual and 3D model containing real-world observations used for training, respectively. This navigation model is chosen to investigate the benefit of employing an unsupervised domain adaptation model during training to reduce the visual domain gap between virtual and real observations, which does not require the reconstruction of a 3D model containing real-world observations as compared to the ``RGB Synthetic + Real'' model, although it still relies on observations from both domains.
\end{itemize}
Pre-training the ``RGB Synthetic + Real'' and the ``RGB Synthetic + CycleGAN'' models on virtual observations resulted in higher performance compared to the same navigation models trained directly on real-world images or transformed images only, as already highlighted in the work of~\cite{rosano2020navigation}.

\subsection{Real-world Evaluation}
To validate the navigation results reported by the proposed realistic evaluation framework based on real observations and, more generally, to assess the ability of the proposed navigation models to operate in a real context without performing any additional domain adaptation between simulation and real world, we have carried out experiments in the office environment using a real robotic platform. We leveraged a robot equipped with accurate sensors and actuators, able to perform precise movements. That is a desired feature to have because imprecise actions could lead to sensible drops in performance~\citep{rosano2020comparison,arewemakingprogress}. Although this is an important issue to address, in this work we focus more on the visual understanding abilities of the navigation models to support the navigation process, thus we defer a further investigation on the impact of noisy sensors and actuation to future works.
We equipped the robot with a Realsense d435i camera which we mounted to match the point of view of the virtual agent.
We set up a client-server communication system to move the computation from the limited hardware of the robot to a more powerful machine.
At each navigation step, the robot takes a RGB image from the real environment and sends it to the server. In the server, the image is processed by the navigation model which returns the action to be executed, that is sent back and executed by the robot. The wheel encoders of the robot provide the system a feedback about the course of motion.

In total, in this experimental setting, we considered six testing navigation trajectories with an increasing level of difficulty to assess the capabilities of the different models to understand the surrounding environment and take the appropriate actions accordingly.
Most episodes require the navigation models to reason about the obstacles that are interposed between the current position of the agent and the goal, and to find the best path given the agent's understanding of the layout of the space, inferred from the current and previous observations collected during the navigation episode. For instance, all goals are out of the line of sight of the agent in the starting pose and, in most of the episodes, the goal is not visible for most of the navigation time; in some episodes the obstacle appears suddenly (i.e. in episode 4 the agent turns around towards the goal and faces the pillar at a very short distance); in other episodes (episodes 5 and 6) a movable obstacle was placed at test time only in order to test the ability of the navigation models to cope with obstacles never seen during training and deal with new space layouts.
All navigation models have been tested on all the real-world trajectories and, to verify the repeatability and the reliability of the learned navigation policies, each navigation episode has been repeated three times.
Testing one model on one episode took about 5 minutes on average, for a total of ~1710 minutes or 28.5 hours required to complete the entire testing procedure.
It should be noted that, in general, evaluating a large number of navigation models in real settings is time-consuming and requires a constant human supervision. Moreover, a lot of influencing factors should be taken into account to minimize the time spent to carry out the task. For instance, aspects such as how slippery or uneven the floor is, the grip properties of the robot's wheels, the failure rate of the robot's actuators and the robot's battery life heavily influence the amount of time needed to perform an extensive performance evaluation task and, in some cases, can totally compromise its execution.
Given the high costs involved with the assessment of the performance on a real robot, it is immediately evident the value offered by the proposed evaluation tool, which drastically reduces the testing time to a few seconds per episode by considering real observations.

\section{Results}
\label{sec:results}
In this section we report the results achieved by the proposed navigation models when evaluated on realistic trajectories in simulation and when tested in the real world using the robotic platform. We then highlight how the proposed evaluation tool can effectively provide valuable performance estimations and show how the proposed navigation models are able to generalize to the real world.
\subsection{Evaluation on Realistic Trajectories}

\begin{figure}[t]
    \centering
    \includegraphics[width=0.99\linewidth]{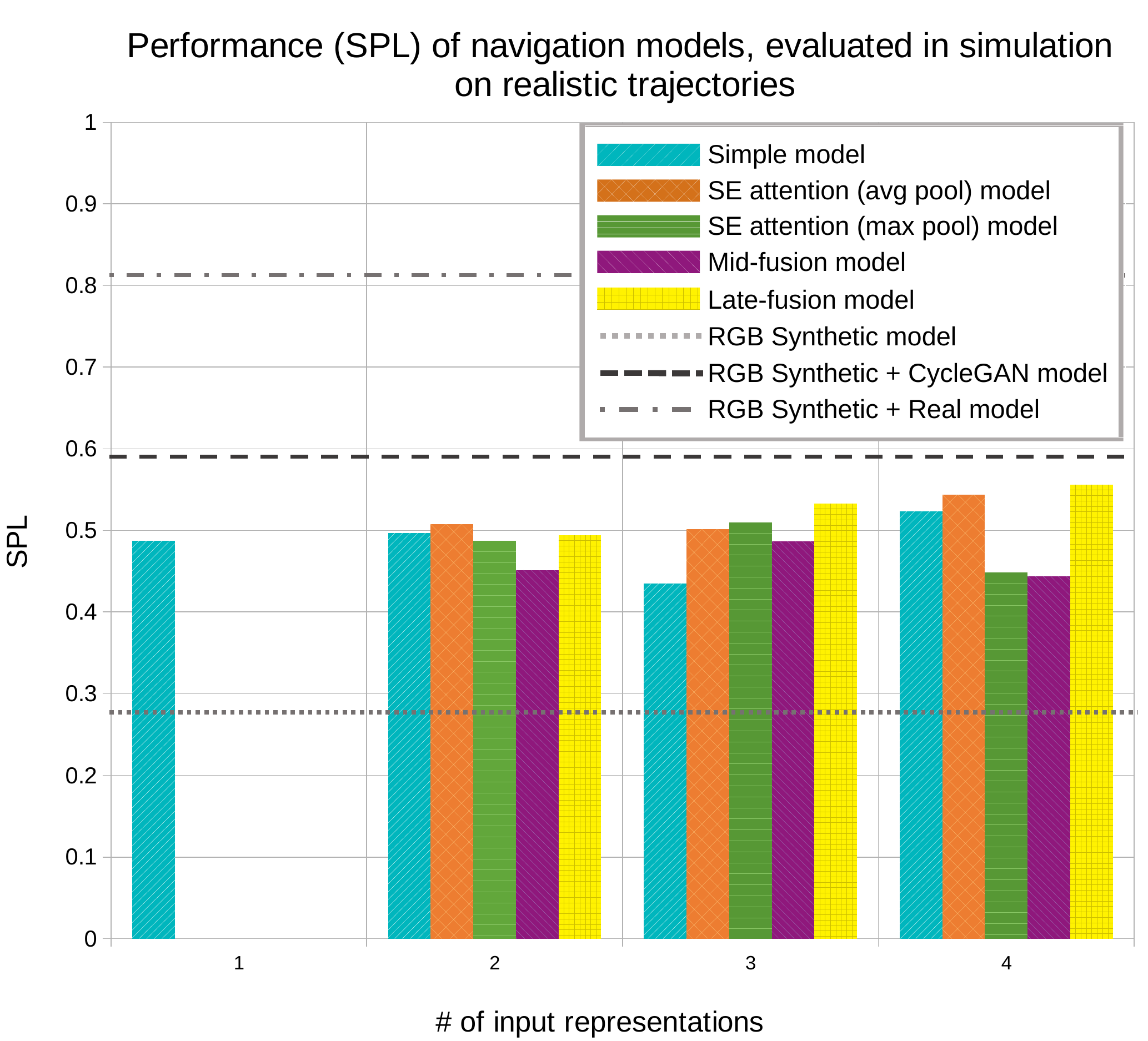}
    \caption{Performance of visual navigation models evaluated in simulation on realistic navigation episodes comprising real-world observations. The performances are reported in terms of SPL (Success weighted by Path Length)}
    \label{figure: SPL_virtual}
\end{figure}

Table~\ref{table:virtual_perf} reports the performance in terms of SPL and SR of all the proposed visual navigation models tested on realistic navigation episodes comprising real-world observations. The notation \textit{A + B} denotes the types and the number of mid-level representations provided as input to the navigation models. As previously highlighted, to increase the number of input visual representations, we followed a greedy approach and expanded the model that reported the best result by adding an additional representation, meanwhile retaining the already used ones. For instance, considering the ``Mid-fusion'' model receiving three representations, we took the best performing ``Mid-fusion'' model with two representations (i.e. surface normals + keypoints3d, ``n + k") and extended it with a third one, curvature or depth, to obtain the ``n + k + c'' and the ``n + k + d'' ``Mid-fusion'' models.
\begin{table}[]
\centering
\begin{adjustbox}{width=0.475\textwidth} 
\begin{tabular}{cccc}
\hline
\rule{0pt}{4ex}
\textbf{
\begin{tabular}[c]{@{}c@{}}Mid-level\\ representations\end{tabular}} & \textbf{Navigation model} & \textbf{SPL} & \textbf{SR} \\ [2ex]
\hline
\rowcolor[HTML]{FFFFFF}
\rule{0pt}{2.5ex}
RGB Synthetic                                                                  & SE-ResNeXt50              & 0.2610       & 0.3990      \\
RGB Synthetic + Real                                                                      & SE-ResNeXt50              & 0.8269       & 0.9640      \\
RGB Synthetic + CG                                                                     & SE-ResNeXt50              & 0.5985       & 0.7500       \\ 
\hline
\rule{0pt}{2.0ex}
(surface) normals (n)                                                                      & Simple model              & 0.4877       & 0.6180      \\
\rowcolor[HTML]{FFFFFF} 
keypoints3d (k)                                                                 & Simple model              & 0.4396       & 0.5740      \\
\rowcolor[HTML]{FFFFFF} 
curvature (c)                                                                   & Simple model              & 0.4262       & 0.5690      \\
\rowcolor[HTML]{FFFFFF} 
depth (d)                                                                       & Simple model              & 0.4417       & 0.5560      \\
\rowcolor[HTML]{E8E8E8} 
n+k                                                                          & Simple model              & 0.4972       & 0.6330      \\
\rowcolor[HTML]{E8E8E8} 
n+c                                                                          & Simple model              & 0.4295       & 0.5530      \\
\rowcolor[HTML]{E8E8E8} 
n+d                                                                          & Simple model              & 0.3515       & 0.4650      \\
\rowcolor[HTML]{DCDCDC} 
n+k+c                                                                        & Simple model              & 0.4349       & 0.5410      \\
\rowcolor[HTML]{DCDCDC} 
n+k+d                                                                        & Simple model              & 0.4314       & 0.5380      \\
\rowcolor[HTML]{CDCDCD} 
n+k+c+d                                                                      & Simple model              & 0.5233       & 0.6630      \\ \hline
\rowcolor[HTML]{E8E8E8} 
n+k                                                                          & SE att. (avg Pool)   & 0.5078       & 0.6400      \\
\rowcolor[HTML]{E8E8E8} 
n+c                                                                          & SE att. (avg Pool)   & 0.4683       & 0.6260      \\
\rowcolor[HTML]{E8E8E8} 
n+d                                                                          & SE att. (avg Pool)   & 0.3847       & 0.5330      \\
\rowcolor[HTML]{DCDCDC} 
n+k+c                                                                        & SE att. (avg Pool)   & 0.5014       & 0.6340      \\
\rowcolor[HTML]{DCDCDC} 
n+k+d                                                                        & SE att. (avg Pool)   & 0.4142       & 0.5490      \\
\rowcolor[HTML]{CDCDCD} 
n+k+c+d                                                                      & SE att. (avg Pool)   & 0.5440       & 0.6840      \\ \hline
\rowcolor[HTML]{E8E8E8} 
n+k                                                                          & SE att. (max Pool)   & 0.4878       & 0.6410      \\
\rowcolor[HTML]{E8E8E8} 
n+c                                                                          & SE att. (max Pool)   & 0.4488       & 0.5897      \\
\rowcolor[HTML]{E8E8E8} 
n+d                                                                          & SE att. (max Pool)   & 0.4585       & 0.6025      \\
\rowcolor[HTML]{DCDCDC} 
n+k+c                                                                        & SE att. (max Pool)   & 0.4943       & 0.6550      \\
\rowcolor[HTML]{DCDCDC} 
n+k+d                                                                        & SE att. (max Pool)   & 0.5101       & 0.6420      \\
\rowcolor[HTML]{CDCDCD} 
n+k+c+d                                                                      & SE att. (max Pool)   & 0.4487       & 0.5730      \\ \hline
\rowcolor[HTML]{E8E8E8} 
n+k                                                                          & Mid-fusion                & 0.4512       & 0.6150      \\
\rowcolor[HTML]{E8E8E8} 
n+c                                                                          & Mid-fusion                & 0.4286       & 0.5846      \\
\rowcolor[HTML]{E8E8E8} 
n+d                                                                          & Mid-fusion                & 0.4128       & 0.5627      \\
\rowcolor[HTML]{DCDCDC} 
n+k+c                                                                        & Mid-fusion                & 0.4851       & 0.6890      \\
\rowcolor[HTML]{DCDCDC} 
n+k+d                                                                        & Mid-fusion                & 0.4870       & 0.6850      \\
\rowcolor[HTML]{CDCDCD} 
n+k+c+d                                                                      & Mid-fusion                & 0.4441       & 0.5910      \\ \hline
\rowcolor[HTML]{E8E8E8} 
n+k                                                                          & Late-fusion               & 0.4944       & 0.6300      \\
\rowcolor[HTML]{E8E8E8} 
n+c                                                                          & Late-fusion               & 0.4845       & 0.6174      \\
\rowcolor[HTML]{E8E8E8} 
n+d                                                                          & Late-fusion               & 0.4573       & 0.5828      \\
\rowcolor[HTML]{DCDCDC} 
n+k+c                                                                        & Late-fusion               & 0.5329       & 0.6790      \\
\rowcolor[HTML]{DCDCDC} 
n+k+d                                                                        & Late-fusion               & 0.5284       & 0.6720      \\
\rowcolor[HTML]{CDCDCD} 
n+k+c+d                                                                      & Late-fusion               & 0.5561       & 0.7110      \\ \hline 
\\ 
\end{tabular}
\end{adjustbox}
\caption{Performance of all the considered visual navigation models, evaluated on realistic navigation episodes in simulation using the proposed evaluation tool. We compare 5 different architectures, each of them receiving between 1 and 4 mid-level representations as input. Together with 3 RGB baselines, a total of 37 navigation models have been trained and tested. The results are reported in terms of SPL (Success weighted by Path Length) and SR (Success Rate) at reaching the navigation goal}
\label{table:virtual_perf}
\end{table}
A summary of the achieved performance is provided in Figure~\ref{figure: SPL_virtual}, which reports the SPL of the best performing model of each pair \textit{\{model type, number of visual representations as input\}}.
First of all, all the proposed models largely outperformed the ``RGB Synthetic'' baseline model, clearly showing the presence of a crucial sim-real domain gap, which was successfully reduced by the adoption of mid-level representations. 
Compared to the ``RGB Synthetic + CycleGAN'' baseline (last row of Table~\ref{table:virtual_perf}), all models reported a slightly lower SPL and SR values, while not requiring any observation of the target domain to perform any adaptation. Indeed, the two best performing mid-level models, namely the ``SE attention (avg pool)'' model and the ``Late-fusion'' model, both with 4 representations as input, achieved an SPL of $0.5440$ and $0.5561$ respectively, against the $0.5985$ of the ``RGB Synthetic + CycleGAN'' model.
As expected, the ``RGB Synthetic + Real'' model benefited from the supervised adaptation procedure, reporting an SPL of $0.8269$ and a near perfect SR of $0.9640$.
Interestingly, the ``Simple'' model reported good performance even in its basic variant with a single visual representation as input, indicating the high capability of mid-level representations to embed relevant properties of the scene that are meaningful for the navigation task. Increasing the number of input representations, we can observe an improvement of the performance with 2 and 4 representations, but a decrease in the case of 3. 
We hypothesize that this inconsistent behavior may be caused by the compact size of the considered model that, together with a very simple representations fusion scheme, could have failed to correctly manage the additional quantity of data received as input.
A similar behavior can be observed with the ``SE attention (max pool)'' model and the ``Mid-fusion'' model, whose results increased with 3 representations and dropped with 4 representations.
In contrast, the ``SE attention (avg pool)'' model reported the same performance when passing from 2 to 3 input representations and showed a significant improvement when trained on 4 representations, overall achieving one of the best results. Moreover, we can observe that it always outperformed the ``Simple'' model receiving the same number of input representations. 
This achievement confirms the effectiveness of the feature-level attention mechanism, meanwhile showing the importance of carefully designing features aggregation scheme, given the performance gap with the similar ``SE attention (max pool)'' model.

Additionally, we noticed that some mid-level representations failed to make a useful contribution to navigation, resulting in limited performance regardless of the fusion model used. This is the case, for example, of the ``depth'' mid-level representation which, when used in models using two and three representations as input, has recorded performances lower than or at most in line with those of the same model using the same number of input representations and, in the case of models using two representations (``n+d'', all models) even lower than those achieved by the best navigation model using one representation. We hypothesize that this unexpected behavior could be due to the redundancy of geometric information captured by the mid-level representations which can be substantial for the ``normals'' and ``depth'' mid-level representations. In such circumstances the information introduced by the additional representation could be unnecessary and the navigation model, in an attempt to exploit all the mid-level representations provided as input, could fail in the process, leading to inferior results.

The ``Late-fusion'' model showed the most consistent behavior, with a performance that increased proportionally with the number of input representations. Late-fusion achieved the best result among models with 3 representations and the absolute best result with 4 representations, with an SPL of $0.5561$ and a SR of $0.7110$.
We believe that the overall model benefited from the specialization of the different branches on given mid-level representations, that had the chance to independently learn the meaningful information to retain from the input representations. With the policy fusion module then, the model had the chance to decide which branch is less or more likely to output an optimal action, given the specific perception.

\subsection{Evaluation in the Real World}

\begin{figure}[t]
    \centering
    \includegraphics[width=0.99\linewidth]{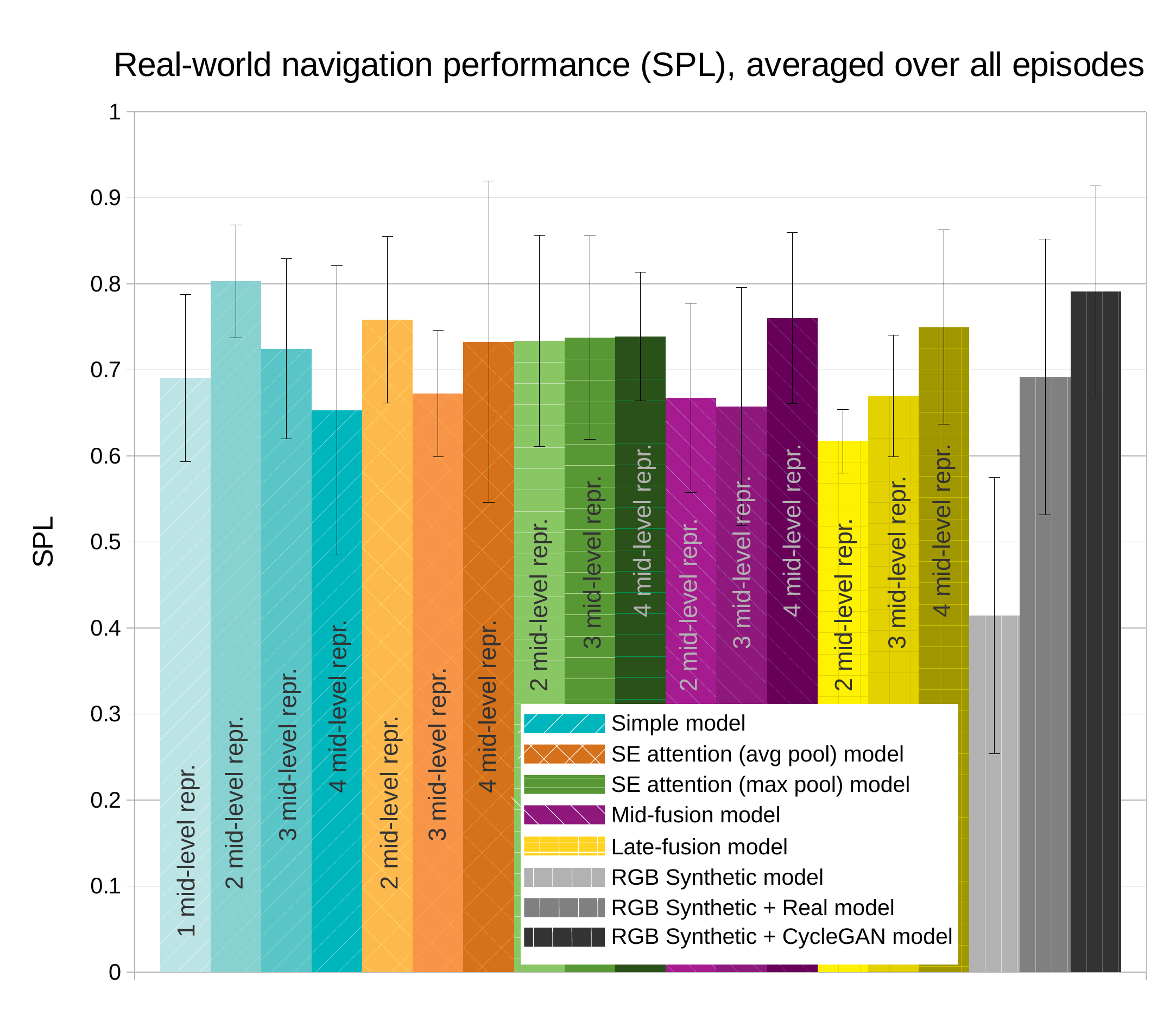}
    \caption{Performance of visual navigation models evaluated in the real-world using the real robotic platform. The performances are reported in terms of SPL (Success weighted by Path Length)}
    \label{figure: SPL_real}
\end{figure}

Figure~\ref{figure: SPL_real} reports the performances obtained by models in the real-world evaluation across all the navigation episodes.
First of all, almost all models successfully reached the navigation goals, with the exception of the ``Simple'' model, which failed in 1 trial out of 18 (the total number of executed trajectories).

Taking a look at the baselines, as expected the ``RGB Synthetic'' model reported the lowest result, confirming that a navigation policy trained in simulation can not be directly transferred to the real world due to the persistency of a sim-real domain shift, that should be addressed with the design of appropriate tools.
Interestingly, the ``RGB Synthetic + CycleGAN'' model returned the best result among the baselines, suggesting that even a general unsupervised domain adaptation technique can effectively help the visual navigation model to address the domain gap.
\begin{figure}[t]
    \centering
    \includegraphics[width=0.99\linewidth]{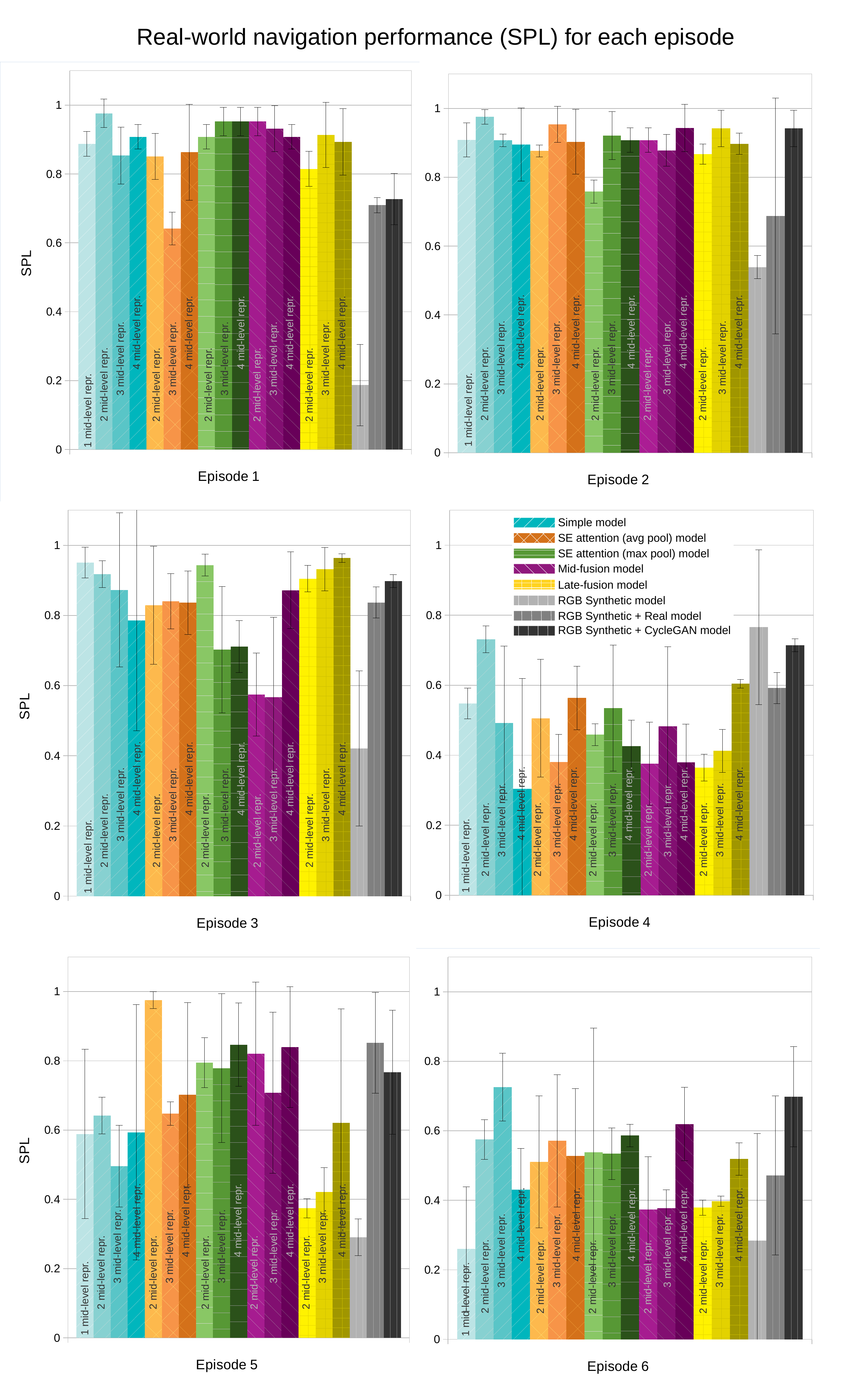}
    \caption{Performance of navigation models in the real world using the real robotic platform, reported for each of the considered trajectories separately. The episodes differ for their complexity, as can be seen from the evolution of the results}
    \label{figure: SPL_real_single_episodes}
\end{figure}

Generally, all the proposed mid-level representation models reported good results, with an average SPL of $0.7103$.
The ``Simple'' model with a single visual representation, it reported a remarkable SPL of $0.6906$, which competes with the result of the ``RGB Synthetic + Real'' baseline. The ``Simple'' model with 2 representations reported an even better result, surpassing the performance of all the baselines, with an SPL of $0.8029$. Although this promising improvement, increasing the number of input representations did not lead to better results. As already hypothesized in the previous subsection, this may be caused by the basic architecture of the model, which limits the scaling of performance with the increase of the input representations.
The ``SE attention (avg pool)'' model achieved interesting performance, with the model trained on 2 representations reporting an SPL of $0.7581$, greater than the models trained on 3 and 4 representations. Also, the ``SE attention (max pool)'' reported similar results, with a maximum SPL of $0.7385$ for the model trained on 4 representations. Both models showed a limited ability to benefit from the additional input, with stable or slightly decreasing performance.
In contrast, a different trend can be observed with the ``Mid-fusion'' and the ``Late-fusion'' models, whose performance increased with the number of input representations, both reaching two peaks of $0.7603$ and $0.7497$ for the ``Mid-fusion'' and the ``Late-fusion'' models with 4 representations respectively. In this case, the models succeeded at exploiting the extra input, positioning themselves among the best performing navigation models.

A more detailed overview of the real-world results is provided by Figure~\ref{figure: SPL_real_single_episodes}, which reports the average SPL values relative to each of the considered real-world trajectories. As expected, in episode 1 most of the navigation models succeeded at following the optimal path, that was short and free of obstacles. Similar results are reported for episode 2, that presents a more challenging scenario but that was successfully managed by most of the proposed models, with very few differences. 
In the episodes from 3 to 6 we can observe a general decrease in performance, given that more sophisticated reasoning ability are required to face the complexity of the trajectories.
The ``Simple'' model with 2 representations performed consistently well in all episodes, outperforming all the proposed models in episodes 1, 2, 4, and still reporting competitive results in the remaining ones.
The ``SE attention (avg pool)'' model with 2 representations excelled in episode 5, the one with the longer trajectory, and performed reasonably well in the rest of the episodes.
Good performances were also achieved by the ``Late-fusion'' model with 4 representations, which showed how using a multi-source input and a more complex architecture can lead to a more stable behavior across a variety of different scenarios, consistently reporting superior or at least comparable performance in each episode relative to the ``Simple'' model with a single representation.

In summary, we can observe that using a spectrum of mid-level representations is a viable approach to designing robust visual navigation models. From the analysis we carried out, no single model has prevailed over the others, but many of them showed to perform well in complex navigation trajectories. Indeed, they achieved better results compared to a classic, single representation model, and their results matched or exceeded the ones of navigation models that had access to observations of the target domain during training (the ``RGB Synthetic + Real'' and the ``RGB Synthetic + CycleGAN'' baselines). 
\begin{figure}[t]
    \centering
    \includegraphics[width=0.99\linewidth]{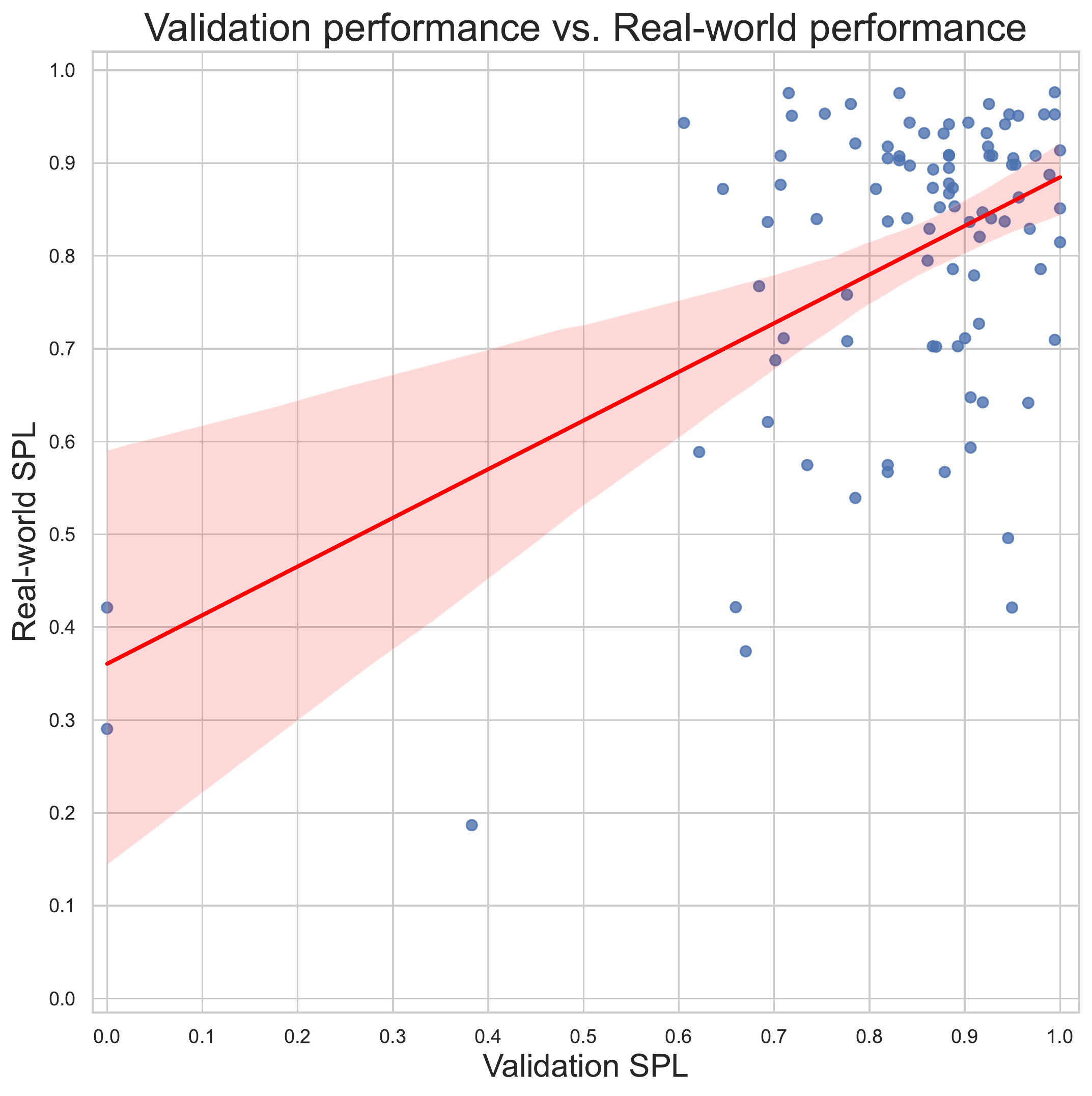}
    \caption{SPL values estimated by our tool in simulation, vs. SPL values measured in the real-world, after performing the same navigation episodes. Each point represents a navigation model, evaluated on a specific trajectory. The red line represents the line of best fit of the data points, while the red shadow represents the confidence interval of Pearson correlation coefficient for a confidence level of $0.95$. Despite the uncertainty, the graph provides a fair indication of the tool's ability to obtain a good picture of the expected real-world navigation performance}
    \label{figure: correlation}
\end{figure}
To measure the ability of the proposed evaluation tool to estimate the expected performance of a visual navigation model, we replicated the proposed real-world navigation episodes inside the Habitat simulator, to then run an evaluation following the same setup of the real world evaluation.
In Figure~\ref{figure: correlation}, it is possible to observe the scatter plot that relates real-world and evaluation performances in terms of SPL, relative to all navigation models. Overall, the proposed tool appears to provide a good estimation of the models performance. 
We conducted a statistical analysis on the relationship between the validation SPL and the real-world SPL, which revealed a Pearson correlation coefficient equal to $0.5164$, with an associated p-value of $8.4396\cdot10^{-8}$, which is below the significance level of $0.05$. The confidence interval of the Pearson correlation coefficient for a confidence level of $0.95$ is equal to $\left [0.3515, 0.6503 \right ]$. This analysis indicates that there is a statistically significant positive correlation between the validation SPL and the real-world SPL. A linear regression analysis between the validation SPL and real-world SPL further reveals very almost zero p-values, well below the significance level of $0.05$, also for the estimation of the intercept and the coefficient associated to the validation SPL, which highlights how the estimation of the position and slope of the regression line shown in Figure~\ref{figure: correlation} is statistically significant. Specifically, we obtained confidence intervals equal to $[0.2880, 0.5720]$ and $\left [ 0.3350, 0.6830 \right ]$ for the intercept and coefficient of the validation SPL respectively. The confidence intervals have been obtained considering a confidence level of $0.95$. Given the analysis above, we believe that, despite the uncertainty, the graph provides a fair indication of the tool's ability to obtain a good picture of the expected real-world navigation performance.
Additionally, we reported the percentage of SPL values correctly estimated by the proposed evaluation tool, varying the accepted estimation error. In particular, the estimation is considered correct if the SPL value measured in the real-world test was at most $x$ points worse. 
We considered this metric to understand if an increment in the estimated performance reflects an increment in real-world performance, still allowing a margin of error.
More than $70\%$ of the estimated performance values were off by at most $0.1$ SPL points, and more than $85\%$ of the estimated performance reported an SPL value that was at most $0.2$ points apart from the real SPL performance. We believe that the obtained results are fairly satisfactory, given the benefit offered by the evaluation tool in terms of saved time and resources, normally required to assess the performance of a navigation policy in the real world with a real robotic platform.

\subsection{Generalization to Unseen Environments}
Additional experiments were conducted to verify the ability of the proposed navigation models to generalize to environments not seen during training. In particular, since we are interested in testing the quality of the navigation policies on real environments, we decided to leverage the already collected realistic 3D model comprising real-world observations and use it as the test environment, then changing the training scenes.
We therefore chose the navigation model that reported the best results (``Late-fusion n+k+c+d'') and trained it on the Gibson dataset scenes~\citep{xiazamirhe2018gibsonenv}. Specifically, we trained a model on the 72 scenes of the Gibson training set split proposed by~\cite{habitat19iccv} (``Late-fusion-72'') and another model trained on only one scene of the Gibson dataset (``Late-fusion-1"), chosen specifically for having similar properties to the proposed office environment. We decided to consider these two scenarios for two reasons: 1) verify the ability of the navigation model to generalize on new environments when a limited number of training environments is adopted, while still using domain-invariant visual representations; 2) verify if scaling up the number of training scenes improves the generalization capability of the model on real environments. The trained models were then evaluated in simulation on the realistic navigation episodes of the proposed environment. The ``Late-fusion-1'' model achieved a low SPL value of $0.0654$, while the ``Late-fusion-72'' model achieved a higher SPL value of $0.2786$. These results show the presence of a significant difference with the results obtained by the navigation models trained and tested on the same environment but using synthetic and real-world observations, respectively. Increasing the number of training scenes has certainly led to an increase in performance, which however remains far from those obtained by the proposed navigation models. The results suggest that to achieve optimal results, such as to allow the deployment of the system in real environments, it is necessary  to train or at least fine-tune the navigation policy on the synthetic version of the specific target environment.

\section{Conclusion}
\label{sec:conclusion}
In this work we investigated the problem of learning robust visual navigation policies in simulation that can be successfully transferred in the real world. To achieve this goal,  
we proposed an evaluation tool, built on top of the Habitat simulator, able to reproduce realistic navigation trajectories in simulation, overcoming the limits posed by the evaluation on a real robot.
The tool has been then employed to evaluate the performance of a set of visual navigation models performing a combination of mid-level representations, after being trained solely in simulation.
Our results suggest that navigation models can benefit from the additional representations provided as input, given the remarkable performances reported by most of the considered navigation models, even when adopting simple representation fusion strategies.
The real-world test with a robotic platform confirmed the effectiveness of the evaluation tool and the concrete possibility to successfully deploy in the real world navigation policies trained in simulation, without performing any domain adaptation.
A further evaluation on new, diverse environments could help provide more insights on the strengths and limits of the proposed navigation models, in order to further improve their capabilities to successfully operate in real-world scenarios.

\section{Declarations}
\textbf{Funding} This research is supported by OrangeDev\footnote{OrangeDev: \href{https://www.orangedev.it/}{https://www.orangedev.it/}} s.r.l., by Next Vision\footnote{Next Vision: \href{https://www.nextvisionlab.it/}{https://www.nextvisionlab.it/}} s.r.l. and the project MEGABIT - PIAno di inCEntivi per la RIcerca di Ateneo 2020/2022 (PIACERI) – linea di intervento 2, DMI - University of Catania.

\noindent
\textbf{Availability of data and materials} The 3D models, the images and the code proposed in this article are publicly available at the following link:~\href{https://iplab.dmi.unict.it/EmbodiedVN/}{https://iplab.dmi.unict.it/EmbodiedVN/}

\bibliography{sn-bibliography}


\end{document}